\documentclass[10pt,twocolumn]{asme2ej}

\usepackage{epsfig} %% for loading postscript figures
\usepackage{todonotes}
\usepackage{amsmath}
\usepackage{graphicx}
\usepackage{multirow}
\usepackage{multicol}
\usepackage{bm}
\usepackage{array}
\usepackage{enumitem}
\usepackage{enumitem}
\usepackage{amssymb}
\usepackage{float}
\usepackage{stfloats}
\usepackage{placeins}

\title{From Safety Standards to Safe Operation with Mobile Robotic Systems Deployment}

\author{Bruno Belzile
    \affiliation{
	Postdoctoral Fellow\\
 	École de technologie supérieure\\
 	Department of Mechanical Engineering\\
    bruno.belzile.1@ens.etsmtl.ca
    }	
}

\author{Tatiana Wanang-Siyapdjie
    \affiliation{ Graduate Student\\
 	École de technologie supérieure\\
 	Department of Construction Engineering\\
    tatiana.wanang-siyapdjie.1@ens.etsmtl.ca
    }
}

\author{Sina Karimi
    \affiliation{ Graduate Student\\
    École de technologie supérieure\\
    Department of Construction Engineering\\
    sina.karimi.1@ens.etsmtl.ca
    }
}

\author{Rafael Gomes Braga
    \affiliation{ Graduate Student\\
    École de technologie supérieure\\
    Department of Mechanical Engineering\\
    rafael.gomes.braga.1@ens.etsmtl.ca
    }
}

\author{Ivanka Iordanova
    \affiliation{ Professor\\
    École de technologie supérieure\\
    Department of Construction Engineering\\
    ivanka.iordanova@etsmtl.ca
    }
}

\author{David St-Onge
    \affiliation{ Professor\\
    École de technologie supérieure\\
    Department of Mechanical Engineering\\
    Email: david.st-onge@etsmtl.ca
    }
}

\begin{document}

\maketitle

%%%%%%%%%%%%%%%%%%%%%%%%%%%%%%%%%%%%%%%%%%%%%%%%%%%%%%%%%%%%%%%%%%%%%%
\begin{abstract}
{\it 
Mobile robotic systems are increasingly used in various work environments to support productivity. However, deploying robots in workplaces crowded by human workers and interacting with them results in safety challenges and concerns, namely robot-worker collisions and worker distractions in hazardous environments. Moreover, the literature on risk assessment as well as the standard specific to mobile platforms is rather limited. In this context, this paper first conducts a review of the relevant standards and methodologies and then proposes a risk assessment for the safe deployment of mobile robots on construction sites. The approach extends relevant existing safety standards to encompass uncovered scenarios. Safety recommendations are made based on the framework, after its validation by field experts. 
%This paper is divided into three sections. First, a review of OSH standards in construction sites and for manufacturing robots is done. Then, the technical framework is detailed. The latter notably consists of preliminary data collection, semi-structured interviews, risk assessment with quantifiable index and the proposal of risk-reducing measures and their effect on that same index. Finally, a case study involving the deployment of a Clearpath Jackal on a real construction site is included.
}
\end{abstract}

% %%%%%%%%%%%%%%%%%%%%%%%%%%%%%%%%%%%%%%%%%%%%%%%%%%%%%%%%%%%%%%%%%%%%%%
% \begin{nomenclature}
% \entry{A}{You may include nomenclature here.}
% \entry{$\alpha$}{There are two arguments for each entry of the nomemclature environment, the symbol and the definition.}
% \end{nomenclature}

% The primary text heading is  boldface and flushed left with the left margin.  The spacing between the  text and the heading is two line spaces.

% %%%%%%%%%%%%%%%%%%%%%%%%%%%%%%%%%%%%%%%%%%%%%%%%%%%%%%%%%%%%%%%%%%%%%%
\section{Introduction}

The safety of robotic systems is a core aspect of most deployments, especially in regard to human-robot collaboration in shared workplaces. The various approaches proposed can be regrouped following the type of locomotion: Unmanned Aerial Vehicles (UAVs) and Unmanned Ground Vehicles (UGVs). For instance, UAV safety has been vastly studied outdoor~\cite{izadi2018measuring,moud2019qualitative}. Still, the direct/indirect common classification of risk for UAV can be transferred to other locomotion: direct hazards occur when the platform collides with obstacles, such as equipment and labor, while indirect hazards are the environmental circumstances affecting the robot functionality, such as noise and wind. A model to assess the safety of mobile robot deployment with respect to its physical characteristics and its environment was proposed by Moud, Hashem Izadi, et al.~\cite{moud2020safety}.

Along the same line, Augustsson et al.~\cite{8542547} proposed the use of defined safety zones around the robot. The safety zone geometries are flexible and adapt to the work carried out by the robot, which can provide a safe human-robot interaction. Truong et al.~\cite{7057375} proposed another approach for safe collaboration between human and robots. They emphasize on human safety while robots and humans are navigating in a shared environment. When individuals are detected through the robot's sensors, the robotic system generates itself a safe zone for navigation. Shin et al.~\cite{8675665} proposed a framework compliant with ISO/TS 15066~\cite{ISO/TS_15066:2016} (see Section~\ref{s:standards}) to assess the \textit{collision peak pressure} of collaborative robots close to humans. Their study emphasis the importance of the safety aspects in a shared work space. However, the risk assessment of collaborative \emph{mobile} robots is not covered. Dohi et al.~\cite{8460869} introduced a novel safety concept for human-robot collaboration in manufacturing industry that considers robot knowledge of the operator skills and states to enhance safety measures. The authors argue that this type of new \textit{Safety 2.0} paradigm is becoming a necessity for any robotic deployment close to the operators.
%In this direction, there is a growing need in the construction industry (with their unique and dynamic characteristics) for a risk assessment of deploying mobile robots. 

The construction industry has been known to be one of the most dangerous for human workers~\cite{Carter2006SafetyProjects} and using robots in construction sites has many foreseeing advantages for tedious and repetitive tasks (site surveys, material transport, etc.). Therefore, we trust this context to be one of the most challenging environments for safety issues with autonomous mobile robots and we select it for our deployment scenario. Kim~et~al.~\cite{Kim2020ProximityConstruction} already stressed that many security concerns regarding the deployment of robots on construction sites remained unanswered, particularly regarding human-robot collisions (direct or indirect injuries), an issue exacerbated with mobiles robots. Early works on human-robot collaborative tasks in construction sites show the increase of efficiency it can provide, but also the complexity of keeping both that level of efficiency and the safety of the operators~\cite{khatib1987unified}. Brosque et al.~\cite{brosque2020human} follow this approach by implementing several collaborative unit tasks, such as bolting and welding, required to assemble a spatial structure in a simulated environment. Moreover, virtual environments were shown to be helpful to identify and mitigate the risks before conducting the real operations~\cite{you2018enhancing}. Their results are used for the development of a framework for safety analysis of human-robot collaboration. Kim et al.~\cite{kim2019semantic} argue that analogue to human vision helping human assess the risk by semantic classification of the work, robots can use computer vision techniques to become safer. In this direction, they develop a deep neural network architecture to detect and classify semantic construction activities for robots. The results can facilitate the implementation of adaptive safety measures. Despite the meaningful contributions of these studies, none provide a complete assessment of the risks for a mobile validated in field deployment (as opposed to simulation).

Moreover, when studying construction health and safety management systems, Gangolells et al.~\cite{Gangolells2013ModelFirms} highlighted that most companies have a unique and distinct structure of risk assessment and mitigation techniques. This situation can lead to duplicated managing tasks when different standards are applied in parallel; a conflicting situation to avoid.

%%%%%%%%%%%%%%%%%%%%%%%%%%%%%%%%%%%%%%%%%%%%%%%%%%%%%%%%%%%%%%%%%%%%%%%%%%%%%%
\begin{table*}[h]
	\caption{List of relevant standards}
	\begin{center}
		\label{t:standards}
    \begin{tabular}{m{1.5cm}||m{5cm}|m{9.4cm}} 
	Standard & Title & Field of application\\
	\hline \hline
	 ISO 10218~\cite{ISO10218}             &  Robots and robotic devices — Safety requirements for industrial robots — Part 1: Robots \& Part 2:  Robot systems and integration            &   This standard specifies requirements and recommendations for intrinsic prevention, protective measures and information for the use of industrial robots. It describes the basic hazards associated with robots and provides the basic requirements for reducing or eliminating the risks associated with these hazards.    \\ \hline
	 ISO/TS 15066   &  Robots and robotic devices — Collaborative robots  & This standard specifies the safety requirements for collaborative industrial robot systems and the working environment, and complements the requirements and guidance on collaborative industrial robot operation given in ISO 10218-1 and ISO 10218-2 .
	 \\ \hline
	 ANSI / RIA R15.08~\cite{ANSI/RIA_R15.08-1-2020} & Industrial mobile robots’ safety & This standard defines the safety requirements for manufacturers of industrial mobile robots part 1; Part 2 describes the requirements for integrators working on the design, installation and integration of a safe mobile robot system in a user's facilities; and part 3 defines the safety requirements for the end user of industrial mobile robots.
	 \\ \hline
	 ANSI /ITSDF B56.5~\cite{ANSI/ITSDF_B56.5-2019}   &  Safety Standard For Guided Industrial Vehicles & This standard defines the safety requirements relating to the elements of design, operation and maintenance, industrial vehicles with automatic guidance without mechanical restraint and unmanned and the system of which the vehicles are part.
	 \\ \hline
	 CSA/Z434-14 (2019)~\cite{CAN/CSA-Z434-14} (Canada)  & Industrial robots and robot systems  & This part of ISO 10218 specifies requirements and guidelines for inherent safety design, protective measures and information for the use of industrial robots. It describes the basic hazards associated with robots and provides requirements to eliminate or adequately reduce the risks associated with these hazards.
	 
		\end{tabular}
	\end{center}
\end{table*}
%%%%%%%%%%%%%%%%%%%%%%%%%%%%%%%%%%%%%%%%%%%%%%%%%%%%%%%%%%%%%%%%%%%%%%%%%%%%%%%%%

More than a threat, robots navigating a construction site can also be leveraged to enhance the workers safety. For instance, Wang et al.~\cite{wang2017application} proposed UAV aerial images to help with offline and offsite analysis of safety issues. Alizadehsalehi et al.~\cite{alizadehsalehi2020effectiveness} studied the use of Building Information Modeling (BIM) with UAV-based surveys to dynamically identify hazards and their locations on the site. They aim at supporting the construction workers with their risk assessment. The use of BIM helps create a more systemic approach to risk assessment, a critical aspect of effective safety regulation on construction sites, as discussed by Hussain et al.~\cite{hussain2017safety}. Hussain et al. applied Occupational Safety and Health Administration (OSHA) guidelines to propose a framework from the design phase to the surveillance of hazards on construction sites. In these studies, the authors did not consider the potential treat pose by the robot itself.

Finally, the robots deployed must also detect threats to themselves. Ibrahim and Golparvar-Fard~\cite{ibrahim20194d} studied safe fight path of an UAV considering the hazards for autonomous robot navigation in outdoors. Karimi et al.~\cite{karimi2021semantic}, proposed an intelligent path planning method, based on BIM, to avoid potential hazardous areas. Although that there are many applications for robots' deployment on construction sites, the literature shows a lack of risk assessment framework.

In this paper, we propose a framework to assess potential safety risks regarding the deployment of mobile robots on construction sites with a quantifiable index. The paper is divided in the following sections: we detail the relevant standards in Section~\ref{s:standards}, followed by the actual proposed framework in Section~\ref{s:framework}. Finally, we conduct a case study on the deployment of a Clearpath Jackal on a real construction site in Section~\ref{s:case} with the purpose to validate the framework, and to be able to define safety recommendations.

%%%%%%%%%%%%%%%%%%%%%%%%%%%%%%%%%%%%%%%%%%%%%%%%%%%%%%%%%%%%%%%%%%%%%%
\section{Relevant Standards}
\label{s:standards}

Several standards must be considered regarding the deployment of mobile robots in workplaces. While none of them are both specific to mobile robotic systems and applicable to various deployment scenarios, many elements apply. Many of the current standards target manufacturing robots, eg., fixed (and recently some industrial mobile robots), collaborative devices, automated guided vehicles (AGV), automated agricultural machines, etc. Table~\ref{t:standards} display a summary of the relevant standards to mobile robotic systems' safety. Only type C standards, as defined by Villani et al.~\cite{villani_survey_2018}, i.e. those focusing on machine safety, are mentioned here. Nevertheless, basic (type A) and generic safety (type B) standards should still be taken into account in any workplace, but are not related to the mobile platforms threats. ISO/TS 15066 is of particular interest, because it focuses on collaborative robots. Its classification of collaborative tasks with regards to the safety requirements can be translated to mobile robots deployed alongside workers. One, or a combination, of the following four collaborative operating modes, are detailed:
\begin{enumerate}
    \item safety-rated monitored stop;
    \item hand guiding;
    \item speed and separation monitoring;
    \item power and force limiting.
\end{enumerate}
The most recent standard, ANSI/RIA R15.08, focuses on industrial mobile robots (IMR) and was released in December 2020. Both autonomous mobile robots (AMR) ang AGVs are considered IMRs, thus covered by this standard. However, AGVs and AMRs have different scopes: the former follows fixed routes, while the latter uses sensors to avoid and go around obstacles. The ANSI/RIA R15.08 is based on relevant guidance from ANSI/RIA R15.06 and ANSI/ITSDF B56.5, which focus on industrial robot safety and guided industrial vehicles, respectively.

In parallel to those standards, we must also consider local regulations that apply to the specific workplace targeted. Indeed, national and local codes and regulations take precedence over industry standards. Health and Safety Standards are obligatory to follow on construction sites. There is a Safety Agent on each site. Examples of such regulations are discussed in the case study in Section~\ref{s:case}.

\section{Ensuring Safe Operation}
\label{s:framework}

The strategy applied here is divided into four steps, i.e., preliminary planning, data collection, risk assessment, and risk mitigation. The terminology used is inspired by the standards considered above.

\subsection{Preliminary Planning and Data Collection}
In the preliminary planning phase, several elements must be identified and taken into account: the legal aspects affected by the deployment of a robot on a construction site, the objectives motivating the planned robotic solution, and management changes resulting from these new working methods. When this administrative process is completed, some information must then be collected in order to carry out a risk analysis. This is notably done with a questionnaire to be answered by those that will be involved with the robot's deployment. Additionally, semistructured interviews with the same individuals and on-site visits complete the set of information. Taking photos and videos is intrinsic to the last step and is particularly useful to identify potential hazards.

\subsection{Risk Assessment}

The risk assessment is divided in three steps:
\begin{enumerate}
    \item Determination of the parameters of the robot and the environment;
    \item Identification of the potential dangerous phenomena;
    \item Risk evaluation.
\end{enumerate}
First of all, the robot parameters must be identified, which include the type of robot, its shape, its functions, the on-board sensors (which could be used for risk mitigation), the battery's parameters, and any other relevant information. The environment's parameters where the robot will be deployed must also be established, i.e., indoor vs outdoor, type and configuration of the ground, available space to move, presence of obstacles, etc. Finally, elements such as visibility, which depends on both the robot and the environment, must also be taken into account. 

Then, the working conditions of the robot are determined. This includes the typical and maximum duration of use, the frequency and the materials with which the IMR will interact. The complete life cycle of the robot must also be considered, i.e. commissioning, use, troubleshooting and servicing. Potential consequences of bad use and reasonably foreseeable failures must also be known. Finally, the minimum experience needed to operate the robot has to be established beforehand. 

Once the operation parameters of the robot are known, we can identify potential dangerous phenomena, which is a critical step since they must be meticulously listed to minimize the risks. These can be mobile and moving parts (mechanical phenomena), live wires (electrical phenomena), too hot/cold machine parts (thermal phenomena), noise/vibrations, visible (laser) and invisible electromagnetic phenomena, dangerous substances and noncompliance with ergonomic principles. These potential hazards are then classified based on the entities facing the potential dangerous phenomenon, namely workers, the robotic platform itself, and the environment.    

\begin{figure*}[htbp]
  \centering
  \includegraphics[width=0.95\textwidth]{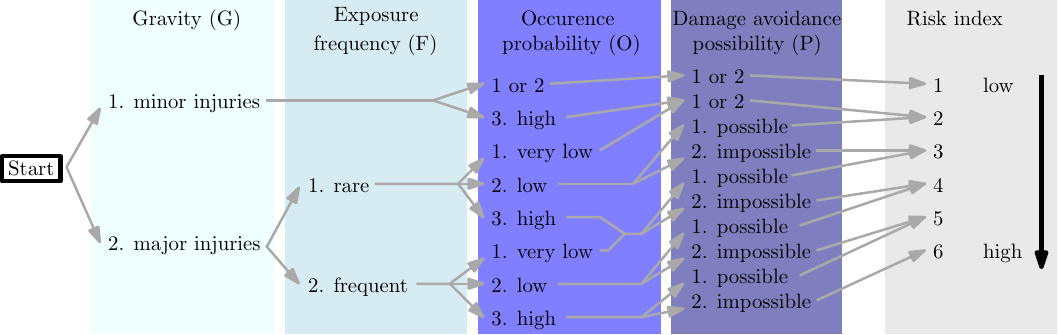}
  \caption{Flow chart of the risk assessment (adapted from \cite{bourbonniere_guide_2005})}
  \label{f:Flowchart}
\end{figure*}

The next step is to evaluate and conduct a relative comparison between the potential risks posed by every dangerous phenomenon identified, in order to establish a priority order for the actions to be taken. The risk is defined as the combination of the probability of damage and its gravity according to ISO 14121. This definition is central to our framework and the risk index used. The latter can be divided into four parts, i.e., (1) gravity/severity of potential injury (G); (2) frequency and exposure time (F); (3) probability of occurrence (O); (4) possibility of avoiding or limiting damage (P). Together, these four parameters form the risk quadruplet set, with one additional parameter compared to the risk triplet set proposed by Prassinos and Lyver~\cite{prassinos_risk_2011}:
\begin{equation}
    RISK = \{<G,F,O,P>\}
    \label{e:quadruplet}
\end{equation}
The flow chart to select a value for these four parameters and the final risk index was designed by Bourbonnière et~al.~\cite{bourbonniere_guide_2005}, and is illustrated in Fig~\ref{f:Flowchart}.

The final stage is to make a judgment on the estimated level of risk, using an index proposed by Giraud~\cite{giraud_securite_2008}. It is at this stage that we determine whether this risk is tolerable or not. When the risk is considered intolerable (high risk index), risk mitigation measures must be chosen and implemented. To ensure that the solutions chosen make it possible to achieve the risk reduction objectives without creating new potential dangers, the risk assessment procedure must be repeated once the solutions have been implemented~\cite{giraud_securite_2008}. Furthermore, this last step is particularly important and cannot be overlooked, as corrective measures to reduce the risk posed by a particular phenomenon can also generated or increase the probability of other dangerous phenomena.

\subsection{Risk Mitigation}
Finally, the last step is minimizing the risk index for all dangerous phenomena identified in the previous section, and potentially eliminating them if possible. Preventive and corrective measures must be put in place and the risk index must then be reevaluated accordingly. These measures are classified into three categories:
\begin{itemize}[label=$\circ$]
    \item intrinsic prevention;
    \item organizational;
    \item guards and detection devices.
\end{itemize}
Examples of intrinsic prevention and mitigation measures at the source are given below:
\begin{itemize}[label=$\square$]
    \item Separation distance between the robot and obstacles (workers, equipment);
    \item Keeping the batteries at a good level of charge or else completely change the battery;
    \item Ensuring a safe form factor of the robot;
    \item Stable ground surface for the robot;
    \item Reducing speed and force/torque capabilities of the robot;
    \item Automatic brake when the robot loses control;
    \item Reducing administrative staff on the site;  
    \item Using technological tools to visualize the construction process before workers access the location;
    \item Using sensors with a range of 2 to 4~m;
    \item Supervision of the workers' movement by a site supervisor.
\end{itemize}
Similarly, a list of organizational-type measures is compiled:
\begin{itemize}[label=$\square$]
    \item Daily safety meetings;
    \item Training of workers;
    \item Compliance with the manufacturer's instructions;
    \item Access roads to the site in order;
    \item Weekly health and safety inspections;
    \item Preventive maintenance of the robot;
    \item Keeping the premises clean, without avoidable obstacles, with an adequate site layout;
    \item Work procedures (safety analysis of tasks, safety breaks);
    \item Preventive inspection of the site before workers' arrival;
    \item Pedestrian traffic plan indicating traffic lanes, road markings;
    \item Work timetable;
    \item Rewarding workers for safe behavior;
    \item Awareness / communication of the robot's operating rules;
    \item Employing a full-time security guard on the site;
    \item The nature of the project (the complexity of the site and the number of workers);
    \item Ensuring individual worker involvement regarding safety measures;
    \item Workers' knowledge of safety measures;
    \item Assigning safety responsibilities to each person on the site;
    \item Communication between management and site workers.
\end{itemize}
Finally, here is a list of mitigation measures related to guards and detection devices:
\begin{itemize}[label=$\square$]
    \item Audible alarms;
    \item Personal Protective Equipment (PPE), wearing high visibility safety clothing;
    \item Emergency stop devices;
    \item Reversing alarms;
    \item Sensitive bumpers for presence detection;
    \item Proximity sensors;
    \item Fall arrest sensors;
    \item Permanent lighting on the construction site allowing easier and safer motion of the robot;
    \item Road signs;
    \item Perimeter fences;
    \item Tactile sensors to detect physical contact;
    \item Surveillance cameras.
\end{itemize}

%%%%%%%%%%%%%%%%%%%%%%%%%%%%%%%%%%%%%%%%%%%%%%%%%%%%%%%%%%%%%%%%%%%%%%

\section{Case Study}
\label{s:case}

\begin{figure}[htbp]
  \centering
  \includegraphics[width=0.45\textwidth]{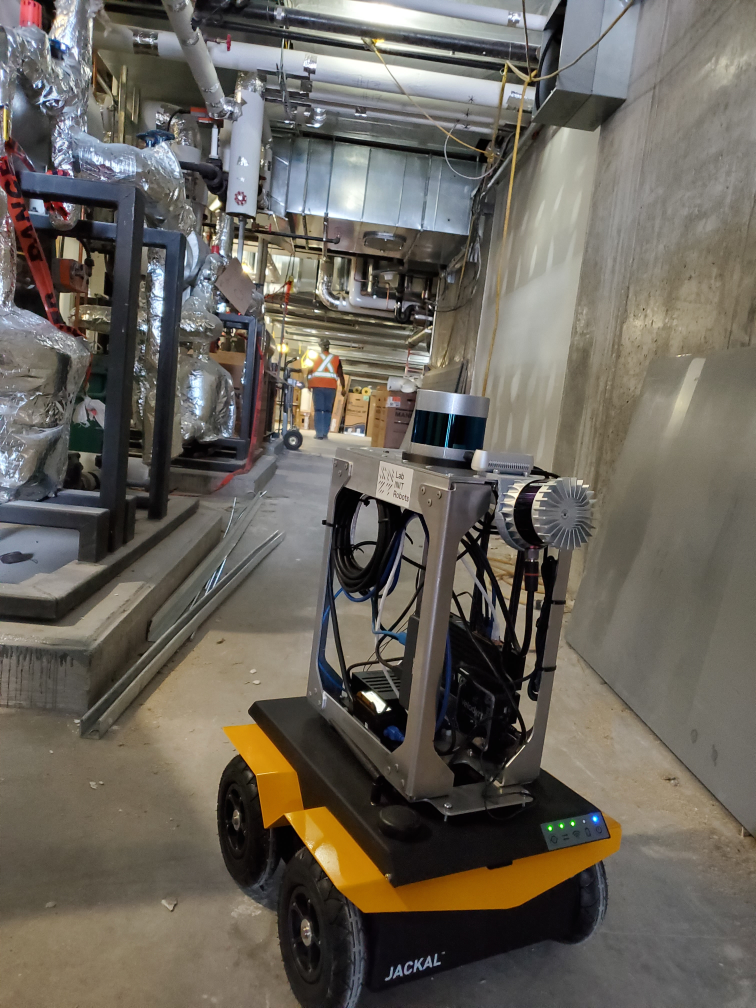}
  \caption{Clearpath Jackal in action on a construction site}
  \label{f:jackal}
\end{figure}

The case study presented here focuses on the intended deployment of a Clearpath Jackal, an autonomous robot depicted on-site in Fig.~\ref{f:jackal}, on a building construction site managed by the company Pomerleau in Montreal, Canada. 

\vspace{-3mm}

%%%%%%%%%%%%%%%%%%%%%%%%%%%%%%%%%%%%%%%%%%%%%%%%%%%%%%%%%%%%%%%%%%%%%

\begin{table*}[h]
	\caption{List of identified potential risks related to the case study with Pomerleau}
	\begin{center}
		\label{t:risks}
    \begin{tabular}{m{0.25cm}||m{2cm}|m{2cm}|m{0.8cm}|m{10cm}} 
	 & Entities facing risk & Primary risk classes & Code & Secondary classes\\
	\hline \hline
	\multirow{4}{0.5cm}{1} & \multirow{7}{2cm}{robot} & \multirow{4}{2cm}{mechanical / physical} & 1-1 & robot incapable of stopping/collision with a worker (speed, force)\\
	 & & & 1-2 & size and shape of the robot \\ 
	 & & & 1-3 & robot overturn (loss of balance)\\ 
	 & & & 1-4 & fall caused by the robot\\ \cline{1-1}\cline{3-5}
	\multirow{2}{0.5cm}{2}  & & \multirow{2}{2cm}{electrical} & 2-1 & contact between various electrical cords in the robot system\\ 
	 & & & 2-2 & low battery \\ \cline{1-1}\cline{3-5}
	3 & & thermal & 3-1 & exposition to extreme temperatures\\ \cline{1-5}
	\multirow{2}{0.5cm}{4} & \multirow{3}{2cm}{worker} & \multirow{2}{2cm}{noise and vibrations} & 4-1 & recurrent false alarms\\ 
	 & & & 4-2 & high ambient noise level of the robot preventing hearing danger alert sound signals or even disrupting conversations between two workers\\  \cline{1-1}\cline{3-5}
	 	5 & & biological & 5-1 & lung infection from a common fungus found in the soil raised by the robot\\ \cline{1-5}
	6 & worker / environment & environment & 6-1 & robot can cause a collision with workers, materials, devices stored or stuck on the ground \\ \cline{1-5}
	\multirow{3}{0.5cm}{7} & \multirow{5}{2cm}{environment} & \multirow{3}{2cm}{environment} & 7-1 & contamination from chemical emissions \\ 
	 & & & 7-2 & environmental conditions (obstacle on the ground preventing the robot and the worker from better deploying\\ 
	 & & & 7-3 & dry and humid weather conditions as well as on hot summer days and frosty winter nights\\\cline{1-1}\cline{3-5}
	 \multirow{2}{0.5cm}{8} & &\multirow{2}{2cm}{substances} & 8-1 & unexpected failure of the robot's mechanical and electrical components\\ 
	 & & & 8-2 & irregular start-up procedure\\\cline{1-5}
	9 & worker / robot / environment & workers related & 9-1 & unqualified person cannot control the robot\\ \cline{1-5}
		\end{tabular}
	\end{center}
\end{table*}

%%%%%%%%%%%%%%%%%%%%%%%%%%%%%%%%%%%%%%%%%%%%%%%%%%%%%%%%%%%%%%%%%%%%%%

\subsection{Local Regulations}
\label{ss:local_regulations}

As mentioned above, beyond the international and national standards, local regulations must also be considered; the deployment of the mobile robot has to comply to the guidelines. In the case at hand, we included the local regulations in Quebec in our analysis, namely the Regulation respecting occupational health and safety ({\it Règlement sur la santé et la sécurité du travail}, RSST) and the Safety Code for the construction industry ({\it Code de sécurité pour les travaux de construction}, CSTC). While these two documents are general and do not explicitly refer to robots, some articles are of interest. We have summarized them in the sequel.

On a construction site, the traffic plan and traffic control are essential, as they constitute a measure to prevent unexpected contact between pedestrians and vehicles. To this end, the project manager must plan the movement of the latter to restrict backing maneuvers and put in place safety measures to protect any person circulating on the site. They must also inform, beforehand, any person who has to circulate on the site of the safety measures provided for in the circulation plan (CSTC, art. 2.8 and 2.8.2). This traffic plan must obviously include any IMR likely to be deployed. The site signaller must meet certain conditions, in particular: wear high-visibility fluorescent yellow-green safety clothing, remain visible to vehicle drivers, and stay out of the latter's path (CSTC, art. 2.8.4). Therefore, any worker must remain visible and identifiable to an IMR by wearing high visibility clothing.

Personal protective equipment must be used where there is a potential contact danger with a mobile part. Naturally, it does not prevent an accident, only mitigates the consequences. These are means of protection, not means of prevention (CSTC, art. 2.10). On any construction site, no danger must result from the storage location and conditions of materials or equipment, which also includes IMRs (CSTC, art. 3.2.1).

Self-propelled vehicles must be equipped with effective brakes and warning devices to be used when approaching pedestrians, doors, turns and dangerous locations. This requirement does not apply to mechanical rams mounted on crawlers, skidders and all terrain vehicles (CSTC, art. 3.10.2). All equipment must only be used by an experienced driver or under their supervision (CSTC, art. 3.10.4). Finally, all self-propelled vehicles must be equipped with an audible warning device which must be installed within reach of the driver, have a sound specific to the type of equipment and different from any other signal from the same construction site and have sufficient noise level to dominate construction noise (CSTC, art. 3.10.12). 

\subsection{Recommendations}

Based on the needs identified, Pomerleau chose the Jackal by Clearpath. Afterwards, we recommended a way forward to deploy it safely. The following recommended steps described are adapted for a worker without any prior experience with robots to be used in a building under construction. 

\textbf{Step 1:} Ensure compliance with robot safety standards ISO 10218-1 and 2 and ANSI / RIA R15.08.

\textbf{Step 2:} Present the Jackal to the various stakeholders (workers, supervisors) on the site in order for them to get to know the new mobile system that will now share their workspace. This step is critical to make sure they know the risks and how to behave in some specific situations with this IMR to reduce the probability of incidents.

\textbf{Step 3:} Make a presentation to be the local safety regulator (CNESST in the case at hand). In this case, Pomerleau has to prove to the CNESST that everything has been considered and validated to guarantee a deployment with the best safety conditions. The CNESST and unions must be aware of the presence of the IMR on the site and if necessary, validate its installation.

\textbf{Step 4:} Mapping the construction site; validating the ground is completely flat and where obstacles are located, which will facilitate the movement of the Jackal. It is essential to verify the data acquisition system on the Jackal with its sensors is functional.

\textbf{Step 5:} Undertaking a risk assessment procedure; This step consists of respecting two essential points: identifying the risks associated with the Jackal, then estimating and evaluating these risks in order to define the priorities.

\textbf{Step 6:} Identifying risk mitigation measures.

\textbf{Step 7:} Validating the risk mitigation measures through on-site simulated scenarios with the Jackal. Identify shortcomings and make adjustments if necessary.

\textbf{Step 8:} Validating the measures put in place while the robot is in operation by tests and feedback from the workers and experts in the field.

\subsection{Analysis and Evaluation}

With the information obtained on the robot and the environment where it will be deployed, the potential risks where compiled and classified, as listed in the table in Table~\ref{t:risks}. A risk index was then estimated with quadruplet of Eq.~\ref{e:quadruplet} for each dangerous phenomenon and was included in the tables in Appendix, with mitigation measures as well. Among the dangers identified, the ones with the highest risk index (6) are:
\begin{itemize}[label=$\circ$]
    \item Surprise effect;
    \item Crushing  of  the  robot by  a  vehicle  or worker;
    \item Unexpected failure of the mechanical   and electrical components   of   the robot (example security device);
    \item Obstacles on the ground preventing the robot from deploying correctly.
\end{itemize}

Finally, based on the risk mitigation measures proposed, a new risk index was estimated for each dangerous phenomenon identified. The results are given in Table~\ref{t:reevaluation}. As can be seen, the risk index of all identified phenomena except \#4 and 12 was reduced to a value of 1, including those with an initial index of 6 mentioned in the list above.  
\begin{table}[h]
	\caption{Residual risk with the Jackal}
	\begin{center}
		\label{t:reevaluation}
    \begin{tabular}{m{1.0cm}||m{0.8cm}|m{0.8cm}|m{0.8cm}|m{0.8cm}||m{0.8cm}} 
    \hline
    Danger & \rotatebox{90}{Severity} & \rotatebox{90}{Frequency} & \rotatebox{90}{Probability} & \rotatebox{90}{Avoidance} & \rotatebox{90}{Risk index}\\
    \hline \hline
    1 & G1 & F2 & O2 & P1 & 1\\ \hline 
    2 & G1 & F2 & O2 & P1 & 1\\ \hline 
    3 & G1 & F1 & O2 & P1 & 1\\ \hline 
    4 & G1 & F1 & O3 & P1 & 2\\ \hline 
    5 & G1 & F1 & O2 & P1 & 1\\ \hline 
    6 & G1 & F1 & O2 & P1 & 1\\ \hline 
    7 & G1 & F1 & O2 & P1 & 1\\ \hline 
    8 & G1 & F1 & O2 & P1 & 1\\ \hline 
    9 & G1 & F1 & O2 & P1 & 1\\ \hline 
    10 & G1 & F1 & O2 & P1 & 1\\ \hline 
    11 & G1 & F1 & O2 & P1 & 1\\ \hline 
    12 & G1 & F1 & O3 & P1 & 2\\ \hline 
    13 & G1 & F1 & O2 & P1 & 1\\ \hline 
    14 & G1 & F1 & O2 & P1 & 1\\ \hline
    15 & G1 & F2 & O2 & P2 & 1\\ \hline 
    16 & G1 & F1 & O1 & P1,2 & 1\\ \hline 
    17 & G1 & F1 & O2 & P1 & 1\\ \hline 
    18 & G1 & F1 & O2 & P1 & 1\\ \hline 
    19 & G1 & F1 & O1 & P1,2 & 1\\ \hline 
    20 & G1 & F1 & O2 & P1 & 1\\ \hline 
    \end{tabular}
	\end{center}
\end{table}
As a further step, the potential dangerous situations could have been modeled with probability distributions, as Hoseyni et~al.~\cite{hoseyni_systematic_2014} proposed, with a response surface fitting to estimate the uncertainties. This will be done in future work.

%%%%%%%%%%%%%%%%%%%%%%%%%%%%%%%%%%%%%%%%%%%%%%%%%%%%%%%%%%%%%%%%%%%%%%
\section{Conclusion}
\label{s:conclusion}

Work environments, including construction sites, are environments with a significant number of safety hazards; the introduction of IMRs to increase productivity creates additional concerns regarding their safety. While a range of standards and regulations exists, there is no clear roadmap for the safe deployment of mobile robots on construction sites. Therefore, in this paper, we proposed a framework to assess the risks and to identify measures of mitigation. The framework was validated through a case study in a real construction site context. Future work is planned for further detailing and risk predictability. 

%%%%%%%%%%%%%%%%%%%%%%%%%%%%%%%%%%%%%%%%%%%%%%%%%%%%%%%%%%%%%%%%%%%%%%
\begin{acknowledgment}
The authors are grateful to the Natural Sciences and Engineering Research Council of Canada for the financial support through its CRD program 543867-2019, to Mitacs for the support of this field study as well as Pomerleau; the industrial partner of the ÉTS Industrial Chair on the Integration of Digital Technology in Construction.
\end{acknowledgment}

%%%%%%%%%%%%%%%%%%%%%%%%%%%%%%%%%%%%%%%%%%%%%%%%%%%%%%%%%%%%%%%%%%%%%%
% The bibliography is stored in an external database file
% in the BibTeX format (file_name.bib).  The bibliography is
% created by the following command and it will appear in this
% position in the document. You may, of course, create your
% own bibliography by using thebibliography environment as in
%
% \begin{thebibliography}{12}
% ...
% \bibitem{itemreference} D. E. Knudsen.
% {\em 1966 World Bnus Almanac.}
% {Permafrost Press, Novosibirsk.}
% ...
% \end{thebibliography}

% Here's where you specify the bibliography style file.
% The full file name for the bibliography style file 
% used for an ASME paper is asmems4.bst.
\bibliographystyle{asmems4}

% Here's where you specify the bibliography database file.
% The full file name of the bibliography database for this
% article is asme2e.bib. The name for your database is up
% to you.

\bibliography{asme2e,references}

% %%%%%%%%%%%%%%%%%%%%%%%%%%%%%%%%%%%%%%%%%%%%%%%%%%%%%%%%%%%%%%%%%%%%%%

\newpage
\appendix       %%% starting appendix
\section*{Appendix A: Jackal Risk Assessment}

\begin{table*}[hb!]
	\caption{Jackal risk analysis}
	\begin{center}
		\label{t:risks1}
    \begin{tabular}{m{3.0cm}|m{3.8cm}|m{1.7cm}||m{0.25cm}|m{0.25cm}|m{0.25cm}|m{0.25cm}|m{0.25cm}||m{3.7cm}} 
    \multicolumn{3}{m{5.5cm}||}{Risk components} & \multicolumn{5}{m{3.4cm}||}{Risk index estimation} & Risk reduction measures \\
	\hline
	Dangers / Dangerous Phenomena & Dangerous Situations & Possible damage & \rotatebox{90}{Severity} & \rotatebox{90}{Frequencies} & \rotatebox{90}{Probability} & \rotatebox{90}{Possibility} & \rotatebox{90}{Risk index} & Preventive and/or corrective measures \\
	\hline \hline
	 1 Collision / Contact during displacement of the robot & Possibility for the worker's feet to approach the robot's wheels & Injuries, shock, stumble & G2 & F2 & 3 & P1 & O5 & Integrate obstacle detection sensors, security distance between the worker and the robot
	 \\ \hline
	 2 Robot shape & Possibility for the worker's feet to get in contact with the robot's vertices (e.g. pointed, sharp) & Injuries, shock & G1 & F2 & 3 & P1 & O5 & Give the robot a safe shape (Manufacturer)
	 \\ \hline
	 3 Robot overturning (loss of balance) & The robot loses balance on the floor or on the stairs & Injuries, shock & G1 & F1 & 3 & P1 & 3 & The state of the floor surface must be adequate for the robot (no openings)
	 \\ \hline
	 4 Objects falling on the robot & The worker can drop their tools on the robot & Injuries, abrasions & G1 & F2 & O3 & P1 & 5 & Provide a protective cage in light metal covering the robot
	 \\ \hline
	 5 Robot's battery low & The worker's hands entering in contact with the battery & Overheating, explosion & G1 & F1 & O1 & P1, 2 & 5 & Charge in advance or change the battery
	 \\ \hline
	 6 Robot's noise level & The worker's ears are exposed to the robot's noises & Effect on the hearing, loss of hearing & G1 & F1 & O2 & P1 & 2 & Reduce / eliminate as much as possible the level of noise
	 \\ \hline
	 7 Surprise effect & The worker can trip & Serious injuries & G1 & F2 & O3 & P2 & 6 & Install an audible warning device and/or a small flag on the robot so that it is visible
	 \\ \hline
	 8 Crushing of the robot by a vehicle or worker & The worker can smash the robot & Spill emissions of contaminants & G1 & F2 & O3 & P2 & 6 & Define traffic lanes for the robot
	 \\ \hline
	 9 Confusion between various cables in the robot system & The worker's hands can approach the internal electric components of the robot & Electric shock, inhalation of toxic fumes & G1 & F1 & O2 & P1 & 4 & Preventive and regular inspections by a qualified person
	 \\ \hline
		\end{tabular}
	\end{center}
\end{table*}

\begin{table*}[h!]
	\caption{Jackal risk analysis (Cont.)}
	\begin{center}
		\label{t:risks2}
    \begin{tabular}{m{3.8cm}|m{3.2cm}|m{2.1cm}||m{0.25cm}|m{0.25cm}|m{0.25cm}|m{0.25cm}|m{0.25cm}||m{3.3cm}} 
    \multicolumn{3}{m{5.5cm}||}{Risk components} & \multicolumn{5}{m{3.4cm}||}{Risk index estimation} & Risk reduction measures \\
	\hline
	Dangers / Dangerous Phenomena & Dangerous Situations & Possible damage & \rotatebox{90}{Severity} & \rotatebox{90}{Frequencies} & \rotatebox{90}{Probability} & \rotatebox{90}{Possibility} & \rotatebox{90}{Risk index} & Preventive and/or corrective measures \\
	\hline \hline
	 10 Unexpected failure of the mechanical and electrical components of the robot (e.g. security device) & Exposure of the worker to dangerous electric phenomena & Poisoning, burn, explosion, electrocution & G1 & F2 & O3 & P2 & 6 & Preventive maintenance
	 \\ \hline
	 11 Inability for the robot to stop during its use & Proximity between the worker's feet and the robot & Overheating, explosion & G1 & F1 & O3 & P1 & 3 & Install an emergency stop button
	 \\ \hline
	 12 Strength and speed of the robot & Possibility of contact with the worker & Robot damaged & G1 & F1 & O2 & P1 & 4 & Decrease the speed of the robot
	 \\ \hline
	 13 Irregular start-up procedure (false alarm) & N/A (Not Applicable) & Explosion & G1 & F1 & O2 & P1 & 2 & Preventive maintenance
	 \\ \hline
	 14 Loading an incorrect program & N/A (Not Applicable) & Explosion & G1 & F1 & O2 & P1 & 2 & Automatic actuation of the brakes
	 \\ \hline
	 15 Obstacles on the ground preventing the robot from deploying correctly & Exposure of the worker to unstable ground & Injuries, power is restored after an interruption, slipping & G1 & F2 & O3 & P2 & 6 & Improve the ground surface to facilitate the robot's access and free up intended paths
	 \\ \hline
	 16 Devices or equipment on the ground & Worker can collide with an equipment on the ground & Light injuries & G1 & F1 & O1 & P1, 2 & 2 & Establish a good layout plan
	 \\ \hline
	 17 Excess dust on the robot & Exposure of the robot to excess dust & Robot fails & G1 & F1 & O3 & P1 & 3 & Remove the robot from the work site and clean it whenever it is covered with dust
	 \\ \hline
	 18 Personnel arriving from all directions & N/A (Not Applicable) & Collision between workers & G1 & F1 & O2 & P1 & 2 & Establish a good pedestrian circulation plan
	 \\ \hline
	 19 Dry and wet weather conditions as well as hot summer days and freezing winter nights & Robot exposure to varying temperatures & Robot is damaged & G1 & F1 & O3 & P1 & 3 & Design of an all-terrain and weather resistant robot
	 \\ \hline
	 20 Unqualified person using the robot & N/A (Not Applicable) & Damage the robot & G1 & F1 & O1 & P1, 2 & 2 & Provide training
	 \\ \hline
		\end{tabular}
	\end{center}
\end{table*}

% %%%%%%%%%%%%%%%%%%%%%%%%%%%%%%%%%%%%%%%%%%%%%%%%%%%%%%%%%%%%%%%%%%%%%%
% \section*{Appendix B: Head of Second Appendix}
% \subsection*{Subsection head in appendix}
% The equation counter is not reset in an appendix and the numbers will
% follow one continual sequence from the beginning of the article to the very end as shown in the following example.
% \begin{equation}
% a = b + c.
% \end{equation}

\end{document}